# Optimal Dynamic Coverage Infrastructure for Large-Scale Fleets of Reconnaissance UAVs

Yaniv Altshuler[1]       Alex (Sandy) Pentland[1]       Shlomo Bekhor[2]

Yoram Shiftan[3]

Alfred Bruckstein[4]

.

[1] MIT Media Lab, {yanival,pentland}mit.edu

[2] Transportation Research Institute, Technion, sbekhor@technion.ac.il

[3] Department of Civil Engineering, Technion, shiftan@technion.ac.il

[4] Computer Science Department, Technion, freddy@cs.technion.ac.il

* This research was supported by the Robert Shillman Fund for Global Security - Technion North-Eastern Partnership.

** This research was supported by the Technion Funds for Security research.

## Abstract

Current state of the art in the field of UAV activation relies solely on human operators for the design and adaptation of the drones flying routes. Furthermore, this is being done today on an individual level (one vehicle per operators), with some exceptions of a handful of new systems, that are comprised of a small number of self-organizing swarms, manually guided by a human operator. Drones-based monitoring is of great importance in variety of civilian domains, such as road safety, homeland security, and even environmental control. In its military aspect, efficiently detecting evading targets by a fleet of unmanned drones has an ever increasing impact on the ability of modern armies to engage in warfare. The latter is true both traditional symmetric conflicts among armies as well as asymmetric ones. Be it a speeding driver, a polluting trailer or a covert convoy, the basic challenge remains the same — how can its detection probability be maximized using as little number of drones as possible. In this work we propose a novel approach for the optimization of large scale swarms of reconnaissance drones — capable of producing on-demand optimal coverage strategies for *any* given search scenario. Given an estimation cost of the threat's potential damages, as well as types of monitoring drones available and their comparative performance, our proposed method generates an analytically provable strategy, stating the optimal number and types of drones to be deployed, in order to cost-efficiently monitor a pre-defined region for targets maneuvering using a given roads networks. We demonstrate our model using a unique dataset of the Israeli transportation network, on which different deployment schemes for drones deployment are evaluated.

# 1   Introduction

Last decade has seen a paradigmatic change in the operational processes of modern armies aerial forces, as designers and commanders have been gradually shifting their focus towards unmanned aerial vehicles (UAVs) [31, 31, 49]. Indeed, over 50% of the planes in the US today are unmanned, with this trend expected to further increase, and be adopted by additional western armies in the coming years [33, 76]. Israel, as a leader of UAV technology and military adoption [35, 67], is currently relying heavily on the use of UAVs in its military force, with such vehicles becoming an increasingly dominant element of its intelligence platforms. This has been the case in visual intelligent (VISINT), and recently also in tactical signal intelligence (SIGINT), which is traditionally in charge of 80% of the information gathered by the intelligence corps [2, 22].

Therefore, as the use of UAVs as an integral component on ongoing intelligence gathering in wartime, as well as during the "battle within the wars" increases, so grows the importance of the need to base this use on an efficient infrastructure. In other words, an innovative small scale dedicated UAV squadron designed for special missions may function perfectly with high redundancy and inefficient use of its resources, but a regular large-scale information gathering that is based on unmanned vehicles operating in swarms, cannot. Furthermore, the lack of an efficient infrastructure that assumes control of the low-level resource utilization tasks means that these tasks must ultimately be taken care of by the human operators (as is being done today) dramatically reducing the number of tasks these can engage, increasing the time it takes them to do so, as well as the overall cost of this process, and ultimately significantly limiting the vehicles operational potential.

As the complexity of the problem increases, so does the impact of optimizing capabilities on the overall resources required in order to guarantee a pre-defined level of performance. In other words, a successful use of large scale swarms of UAVs as a combat and intelligence gathering tool necessitates the development of an efficient mechanism for optimization of their utilization, specifically in the design and maintenance of their patrolling routes.

This work proposes an efficient and robust analytic infrastructure for the deployment of collaborative drone swarms, focusing on its application for tactic intelligence gathering. Specifically, we present an analytic model for devising an optimal reconnaissance strategy for any given threat scenario, defined as : (a) The correlation function between the cost of a single drone and its detection performance; (b) The deployment method used (represented as a monotonically increasing function that models the coverage percentage as a function of the number of units); and (c) The estimated expected cost of an undetected threat.

In other words, for any deployment method of the drones swarm (three of which are discussed in this paper), and an answer to the question "how much detection performance do I get by using a more expensive units" the model generates an optimal strategy that is *threat-specific*, namely — economically optimizing the type and quantity of drones with respect to the cost of an undetected threat.

The proposed technique enables swarms of semi-autonomous vehicles to perform an efficient ongoing dynamic patrolling and scanning of the entire roads and transportation infrastructure in a pre-defined "search region", in a robust and near-optimal way, while guaranteeing detection of targets that are traveling in that region.

We demonstrate the applicability of our model using a comprehensive roads network of the Israeli roads and highways system, containing over 15,000 directed links. The rest of the paper is organized as follows : related work is presented in Section 2. The problem and the proposed patrolling model is presented in Section 3. Analysis is presented in Sections 4 and 5, containing among others an analytic estimation of the required number of drones for a given threat scenario (*Theorem 4.2*), as well as an analytical estimation of the optimal types of drones to be used, if several drones of different costs and performances are available (*Theorem 5.2*). Section 6 presents a theoretical analysis of a case-study of the proposed monitoring method. Section 7 presents a second case study, using a real world transportation network dataset in order to discuss several deployment strategies for drones swarms to be used for different types of search regions. The data used for this analysis is presented and discussed in Section 8. Concluding remarks appear in Section 9. Readers interested in further expansion on the mathematical analysis of this problem and similar ones are invited to read [18] and [19], which also partially overlap with this work.

## 2 Related Work

The need to efficiently monitor transportation networks has been the topic of many studies. For example, in [53], the optimal deployment of air quality monitoring units is discussed. An efficient placement of security monitors can be found in [65]. In land transportation, many works had focused on monitoring the transportation of hazardous materials. For example, infectious disease outbreaks pose a critical threat to public health and national security [26, 36]. Utilizing today's expanded trade and travel, infectious agents can be distributed easily within and across country borders as part of a biological terror attack, resulting in potentially significant loss of life, major economic crises, and political instability. Such threats stress even more the importance of a reliable and efficient transportation monitoring infrastructure. A survey of homeland security related threats and risks regarding transportation infrastructure can be found in [32]. In [61] the trade-off between accuracy and coverage, for given limited resources of sensor devices is discussed. Lam and Lo [59] proposed a heuristic approach to select locations for traffic volume count sensors in a roadway network. [91] proposed a sensor deployment framework to maximize such utilities. This framework has been extended to accommodate turning traffic information [27], existing installations and O-D information content [39], screen line problem [90], time-varying network flows [43] and unobserved link flow estimation [52].

In its military aspects and its relevant scientific literature, it is interesting to mention the roots of this field, dating back to World War II [64, 83]. The first planar search problem considered is the patrol of a corridor between parallel borders separated by a constant width. This problem aimed for determining optimal patrol strategies for aircraft searching for ships in a channel [56], with a more generic theory of optimal scanning that was later proposed in [66].

In recent years this problem had gained popularity with the emphasis on large scale swarms of drones [10, 14–16, 24, 54, 57, 66, 78, 84, 85], using a variety of methods ranging from the analysis of the geometrical properties of the search space [8, 9], the use of multi-agent robotics approaches [50, 79, 86] or [21, 48, 63] for biology inspired designs (behavior based control models, flocking and dispersing models and predator-prey approaches, respectively), economics inspired designs [46, 72, 82, 88] or physics inspired approaches [34, 55] (see [3] for a survey of search and evasion strategies).

In general, most of the techniques used for the task of a distributed monitoring use some sort of cellular decomposition. For example, in [74] the area to be monitored is divided between "monitoring agents" based on their relative locations. In [30] a different decomposition method is being used, which is analytically shown to guarantee a complete coverage of the area. Another interesting work is presented in [1], discussing two methods for cooperative monitoring (one probabilistic and the other based on an exact cellular decomposition).

While in most works the targets of the search mission were assumed to be idle, recent works considered dynamic targets, meaning — targets which after being detected by the searching robots, respond by performing various evasive maneuvers intended to prevent their interception. Some of these works can be seen in [6, 11, 13], whereas a variant that emphasizes the stochastic aspects of the problem can be found in [73].

# 3    Patrolling System Optimizing — Problem Definitions

This work analyzes the design of efficient monitoring systems through its economic perspective. Namely, when a multitude of threat scenarios are available, the main challenge is mitigating these scenarios in the cost-effective way. That is, resolving the real-time uncertainty regarding the exact kinds of threats through the optimization of its overall costs (both the direct costs of the system, as well as the indirect costs of undetected threats). It is therefore crucial to provide planners and operators with a monitoring system that can be reconfigured in real-time, having drones deployed / activated gradually as they are needed (due to new information regarding the unfolding situation, and taking into account other operational requirements or budget constraints).

Our proposed model does not assume any constraint on the deployment scheme itself, as the latter can be highly influenced by a large variety of considerations. Rather, it enables planners to model any given deployment scheme they desire, providing the optimal number and types of drones for each one. Our proposed model also offers operators high flexibility regarding the types of drones to be used — modeled using a trade off between the number of monitoring drones and their detection quality with respect to the number of potential targets each drone can monitor simultaneously (to be denoted as the drones' "*Sampling*" quality, a number ranging between 0 and 1). In this work we assume that a higher sampling quality implies a higher cost per drone. Therefore, the overall cost of the monitoring system can be modeled as :

$$Overalll\ Cost = Cost\ per\ Unit \times Number\ of\ Units$$

whereas the overall monitoring performance of the system can be modeled as :

$$Overalll\ Performance = Monitoring\ Coverage\ Percentage \times Sampling$$

We note that the quality of monitoring units (i.e. their sampling rate) has in fact a *non linear* effect on the overall monitoring quality. For example, assuming the sampling rate of each monitor is $\frac{1}{2}$ then will be able to detect slightly more than half of the vehicles traveling in the network, because some of them might pass by more than one monitor and thus have a chance of being detected by either one. Specifically, such vehicles will have probability of $1 - (0.5)^2 = 0.75$ to be detected. Since drones' deployments is likely be done is much smaller numbers compared to the overall possible locations, it is unlikely that many potential targets would pass by two drones during a single trip. We can therefore use a linear approximation.

**Definition 3.1.** *Let $\mathcal{C}_{ATTACK}(\chi)$ denote the expected cost associated with the damage occurred by an attack $\chi$, used to model an undetected target.*

The estimation of the damage costs are beyond the scope of this paper.

**Definition 3.2.** *Let $M(x) \in [0, 1]$ be a monotonically increasing function denoting the percentage of potential targets moving on the roads network, that is detected and analyzed using $x$ monitoring units, using some given monitoring deployment scheme.*

A monitoring system of $n$ units would therefore be cost-effective as long as :

$$\mathcal{C}_{ATTACK} \geq \frac{n}{M(n)} \cdot Cost\ per\ Unit \tag{1}$$

For modeling the function $M(x)$ we propose to use the well-known *Gompertz function* [47], whose general form is :

$$y(t) = ae^{be^{ct}}$$

The *Gompertz function* is widely used for modeling a great variety of processes, (due to the flexibly way it can be controlled using the parameters $a$, $b$ and $c$), such as mobile phone uptake [75], population in a confined space [40], or growth of tumors [38] (see an illustration in Figure 1). Its ability to model the progress of optimization process as a function of the available resources can also be seen in [4, 5, 7, 17].

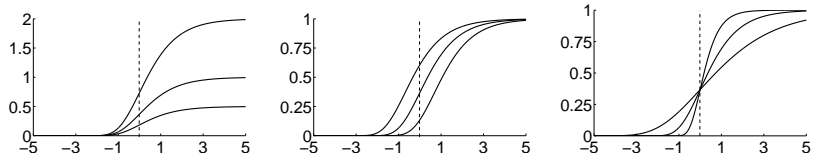

Figure 1: An illustration of the *Gompertz function*. The charts represent the following functions (from left to right) : $y = xe^{-e^t}$, $y = e^{-xe^t}$ and $y = e^{e^{-xt}}$, for $x = \frac{1}{2}$, $x = 1$ and $x = 2$.

The function $M(x)$ can be extrapolated using simulations, as demonstrated in Figure 13. Note that $M(x)$ is domain dependant and may significantly change for different networks.

In the next sections we show how this model can be used in order to calculate the optimal number and types of the monitoring drones.

## 4   The Optimal Number of Monitoring Units

**Definition 4.1.** *The* Normalized Benefit *of a monitoring system of $n$ units is defined as the expected monetary saving from preventing attacks from which the fixed cost of the monitoring units is subtracted :*

$$\omega \triangleq \mathcal{C}_{ATTACK} \cdot M(n) - n \cdot \text{Cost per Unit}$$

Note that this function can refer also to non-monetary aspects of a successful attack. It also resembles the "Net Present Value" calculation, but without explicitly accounting for discount rates.

**Theorem 4.2.** *For any values of $\mathcal{C}_{ATTACK}$, $a$, $b$, $c$ and* Cost per Unit*, the optimal number of monitoring units that would maximize the Normalized Benefit of a monitoring system is :*

$$n_1 = \frac{\ln\left(\frac{1}{a \cdot b \cdot c} \cdot \frac{\text{Cost per Unit}}{\mathcal{C}_{ATTACK}}\right) - W\left(\frac{1}{a \cdot c} \cdot \frac{\text{Cost per Unit}}{\mathcal{C}_{ATTACK}}\right)}{c}$$

$$n_2 = \frac{\ln\left(\frac{1}{a \cdot b \cdot c} \cdot \frac{\text{Cost per Unit}}{\mathcal{C}_{ATTACK}}\right) - W_{-1}\left(\frac{1}{a \cdot c} \cdot \frac{\text{Cost per Unit}}{\mathcal{C}_{ATTACK}}\right)}{c}$$

*where $W(x)$ is the* Lambert product log*, that can be calculated using the series :*

$$W(x) = \sum_{n=1}^{\infty} \frac{(-1)^{n-1} n^{n-2}}{(n-1)!} x^n$$

*Proof.* The optimal value of the Normalized Benefit is received for the number of monitoring units that nullifies the derivative $\frac{\partial \omega}{\partial n}$ :

$$\frac{\partial \omega}{\partial n} = \mathcal{C}_{ATTACK} \cdot \frac{\partial M(n)}{\partial n} - \text{Cost per Unit}$$

Namely :

$$\frac{\partial M(n)}{\partial n} = \frac{\text{Cost per Unit}}{\mathcal{C}_{ATTACK}} \tag{2}$$

(notice that we disregard interest rates, as both cost and damage can assumed to be subject to a similar change along time).

In that case, using Equation 2 we obtain :

$$a \cdot b \cdot c \cdot e^{cn} \cdot e^{be^{cn}} = \frac{\text{Cost per Unit}}{\mathcal{C}_{ATTACK}}$$

which in turn implies :

$$be^{cn} + cn - \ln \frac{\text{Cost per Unit}}{a \cdot b \cdot c \cdot \mathcal{C}_{ATTACK}} = 0 \tag{3}$$

We note that $a > 0$ whereas $b, c < 0$. Analyzing Equation 3 we can then see that in cases where :

$$\frac{Cost\ per\ Unit}{\mathcal{C}_{ATTACK}} \leq -\frac{a \cdot c}{e} \tag{4}$$

the optimal value of $n$ would equal :

$$
\begin{aligned}
n_1 &= \frac{\ln\left(\frac{1}{a \cdot b \cdot c} \cdot \frac{Cost\ per\ Unit}{\mathcal{C}_{ATTACK}}\right) - W\left(\frac{1}{a \cdot c} \cdot \frac{Cost\ per\ Unit}{\mathcal{C}_{ATTACK}}\right)}{c} \\
n_2 &= \frac{\ln\left(\frac{1}{a \cdot b \cdot c} \cdot \frac{Cost\ per\ Unit}{\mathcal{C}_{ATTACK}}\right) - W_{-1}\left(\frac{1}{a \cdot c} \cdot \frac{Cost\ per\ Unit}{\mathcal{C}_{ATTACK}}\right)}{c}
\end{aligned} \tag{5}
$$

where $W(x)$ is the *Lambert product log*, and where $W_k(x)$ is its analytic continuation over the complex plane (the values of the functions $W(x)$ and $W_{-1}(x)$ in the segment implied by the constraint of Equation 4 are illustrated in Figure 2). □

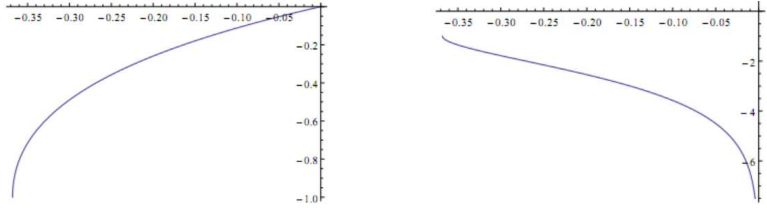

Figure 2: The left and right charts depict the values of the *Lambert* functions $W(x)$ and $W_{-1}(x)$ in the segment $[-\frac{1}{e}, 0]$, respectively. The segment is implied by the constraint of Equation 4.

Returning to the Normalized Benefit of the system, we can now assign the values of the optimal number of monitoring units that appear in Equation 5 into the definition of $\omega$, as follows :

$$\omega_1 = \mathcal{C}_{ATTACK} \cdot a e^{b e^{c n_1}} - n_1 \cdot Cost\ per\ Unit$$

$$\omega_2 = \mathcal{C}_{ATTACK} \cdot a e^{b e^{c n_2}} - n_2 \cdot Cost\ per\ Unit$$

and using the properties of the $W$ function, simplify it into the following form :

$$
\begin{aligned}
\omega_{max} = \max\{\omega_1, \omega_2\} \quad & where: \\
\omega_1 &= a \cdot b \cdot \mathcal{C}_{ATTACK} \cdot \gamma \cdot \left(W\left(b \cdot \gamma\right) + \frac{1}{W\left(b \cdot \gamma\right)} - \ln\left(\gamma\right)\right) \\
\omega_2 &= a \cdot b \cdot \mathcal{C}_{ATTACK} \cdot \gamma \cdot \left(W_{-1}\left(b \cdot \gamma\right) + \frac{1}{W_{-1}\left(b \cdot \gamma\right)} - \ln\left(\gamma\right)\right)
\end{aligned} \tag{6}
$$

where the *Monitoring Benefit Factor* $\gamma$ is defined as:

$$\gamma = \frac{1}{a \cdot b \cdot c} \cdot \frac{Cost\ per\ Unit}{\mathcal{C}_{ATTACK}}$$

From Equation 6 we can see that the Normalized Benefit of a drones monitoring swarm is a function of the *Monitoring Benefit Factor* $\gamma$, which takes into account both the parameters of the threat (through the overall potential damages and the costs of the monitoring units required to detect it) as well as the properties of the monitoring method (characterized by the values of $a$, $b$ and $c$). Notice that the value of *Cost per Unit* does affect the values of $a$, $b$ and $c$ — as they are solely derived from the coverage efficiency of the network's traffic. However, we also note that the actual detection of the drones swarm is affected by their sampling quality, which in turn is monotonously affected by their cost.

## 5    Which Drones to Use? Optimizing the Drones' Cost

In this section we address the issue of finding the optimal type of monitoring drones that should be deployed. This is done by optimizing Equation 6 with respect to the cost per single unit.

**Definition 5.1.** *Let* $\text{Cost}_{\text{Base}}$ *denote the cost of the "most expensive" drone model. That is, the monitoring drone model with the highest level of detection available.*

There is a wide range of monitoring drone models, where in most cases the cheapest ones are expected to have the poorest performance. We define the correlation between the sampling (or detection) quality and the cost of a single unit using a generic function, as follows :

$$Sampling = f_S \left( \frac{Cost\ Per\ Unit}{Cost_{Base}} \right) \tag{7}$$

**Theorem 5.2.** *For any values of* $\mathcal{C}_{ATTACK}$, $a$, $b$, $c$, $\text{Cost per Unit}$, $\text{Cost}_{\text{Base}}$ *and for every function* $f_S : [0, 1] \to [0, 1]$, *the optimal cost for a monitoring unit that would maximize the Normalized Benefit of a monitoring system is a value that satisfies at least one of the following expressions :*

$$W(b \cdot \gamma) - \ln(\gamma) + \frac{1 - \frac{\text{Cost Per Unit}}{\text{Cost}_{\text{Base}}} \cdot \frac{\partial f_S}{\partial \text{Cost Per Unit}} \left[ \frac{\text{Cost Per Unit}}{\text{Cost}_{\text{Base}}} \right] \cdot \left( f_S \left[ \frac{\text{Cost Per Unit}}{\text{Cost}_{\text{Base}}} \right] \right)^{-1}}{W(b \cdot \gamma)} = 0$$

$$W_{-1}(b \cdot \gamma) - \ln(\gamma) + \frac{1 - \frac{\text{Cost Per Unit}}{\text{Cost}_{\text{Base}}} \cdot \frac{\partial f_S}{\partial \text{Cost Per Unit}} \left[ \frac{\text{Cost Per Unit}}{\text{Cost}_{\text{Base}}} \right] \cdot \left( f_S \left[ \frac{\text{Cost Per Unit}}{\text{Cost}_{\text{Base}}} \right] \right)^{-1}}{W_{-1}(b \cdot \gamma)} = 0$$

*where :*

$$\gamma = \frac{1}{a \cdot b \cdot c} \cdot \frac{\text{Cost per Unit}}{\mathcal{C}_{ATTACK}} \cdot \left( f_S \left[ \frac{\text{Cost per Unit}}{\text{Cost}_{\text{Base}}} \right] \right)^{-1}$$

*Proof.* We first revise Equation 6 in order to take into account the different types of monitoring units. as follows :

$$\omega_{max} = \max\{\omega_1, \omega_2\} \tag{8}$$

where $\omega_1$ and $\omega_2$ equal :

$$\omega_1 = a \cdot b \cdot \mathcal{C}_{ATTACK} \cdot Sampling \cdot \gamma \cdot \left( W\left(b \cdot \gamma\right) + \frac{1}{W\left(b \cdot \gamma\right)} - \ln\left(\gamma\right) \right)$$

$$\omega_2 = a \cdot b \cdot \mathcal{C}_{ATTACK} \cdot Sampling \cdot \gamma \cdot \left( W_{-1}\left(b \cdot \gamma\right) + \frac{1}{W_{-1}\left(b \cdot \gamma\right)} - \ln\left(\gamma\right) \right)$$

and where:

$$\gamma = \frac{1}{a \cdot b \cdot c} \cdot \frac{Cost\ per\ Unit}{\mathcal{C}_{ATTACK} \cdot Sampling}$$

We now proceed to the maximizing the financial merits of the monitoring system (namely, $\max\{\omega_1, \omega_2\}$). For this, we shall calculate the partial derivatives $\frac{\partial \omega_1}{\partial Cost\ Per\ Unit}$ and $\frac{\partial \omega_2}{\partial Cost\ Per\ Unit}$ :

$$\frac{\partial \omega_1}{\partial Cost\ Per\ Unit} = \frac{1}{c} \cdot \left( W(b \cdot \gamma) - \ln(\gamma) + \frac{1 - \frac{\partial f_S}{\partial Cost\ Per\ Unit} \cdot \frac{Cost\ Per\ Unit}{Cost_{Base} \cdot f_S}}{W(b \cdot \gamma)} \right) \quad (9)$$

$$\frac{\partial \omega_2}{\partial Cost\ Per\ Unit} = \frac{1}{c} \cdot \left( W_{-1}(b \cdot \gamma) - \ln(\gamma) + \frac{1 - \frac{\partial f_S}{\partial Cost\ Per\ Unit} \cdot \frac{Cost\ Per\ Unit}{Cost_{Base} \cdot f_S}}{W_{-1}(b \cdot \gamma)} \right)$$

and the rest is implied. $\qquad\qquad\square$

Theorem 5.2 can now be used in order to calculate the *optimal aspired cost* of a single drone, for every correlation between its cost and its quality of detection, and for every deployment scheme and estimated threat's potential.

# 6    Case Study I – Theoretical Analysis

In this section we demonstrate how the proposed model can be used in order to produce the optimal number and types of monitoring drones, for a given threat scenario and drones deployment scheme, selected by the system's operators.

For the sake of simplicity we assume in this analysis that both sampling rates and Cost Per Unit to be continuous. This simplifies cases where the sampling rate of the monitoring units can be tuned resulting in lower resource utilization for lower sampling rates (e.g. $f_S(\frac{Cost\ Per\ Unit}{Cost_{Base}}) = \left(\frac{Cost\ Per\ Unit}{Cost_{Base}}\right)^2$). In addition, we assume that the deployment scheme chosen can be modeled with a *Gompertz* function of $a = 1$, $b = -0.2$ and $c = -0.05$ (see an illustration of this function in Figure 3). In this case, nullifying the partial derivative of Equation 9 would produce :

$$\frac{\partial \omega_1}{\partial Cost\ Per\ Unit} = 0 \quad \longrightarrow \quad W\left(-0.2 \cdot \gamma\right) = \ln\left(\gamma\right) + \frac{1}{W(-0.2 \cdot \gamma)} \quad (10)$$

$$\frac{\partial \omega_2}{\partial Cost\ Per\ Unit} = 0 \quad \longrightarrow \quad W_{-1}\left(-0.2 \cdot \gamma\right) = \ln\left(\gamma\right) + \frac{1}{W_{-1}(-0.2 \cdot \gamma)}$$

subsequently implying :

$$\gamma_{opt} \approx 1.77356$$

(in this example, the optimal value of $\gamma$ for $\omega_2$ has a non-zero imaginary component).

Using this optimal value of $\gamma$ we would now get :

$$\gamma_{opt} = \frac{1}{a \cdot b \cdot c} \cdot \frac{Cost\ per\ Unit}{\mathcal{C}_{ATTACK} \cdot Sampling} = \frac{100 \cdot Cost_{Base}^2}{\mathcal{C}_{ATTACK} \cdot Cost\ per\ Unit} = 1.77356$$

and from this we receive :

$$Cost\ per\ Unit_{opt} = \frac{100 \cdot Cost_{Base}^2}{1.77356 \cdot \mathcal{C}_{ATTACK}} \approx 56.384 \cdot \frac{Cost_{Base}^2}{\mathcal{C}_{ATTACK}} \qquad (11)$$

see an illustration of Equation 11 in Figure 5.

From Equation 11 we can obtain for each kind of threat the optimal type of units that should be used, in order to maximize the Normalized Benefit of the system. In this example we also see a linear connection between the optimal cost of the drones and the overall estimations of the damages from a threat.

Assigning this back into Equation 5, we can get the optimal number of units, for each potential threat (see an illustration in Figure 5) :

$$
\begin{aligned}
n_1 &= 20 \cdot W\left(-1127.68 \cdot \frac{Cost_{Base}^2}{\mathcal{C}_{ATTACK}^2}\right) - 172.747 - 40 \cdot \ln\left(\frac{Cost_{Base}}{\mathcal{C}_{ATTACK}}\right) \\
n_1 &= 20 \cdot W\left(-1, -1127.68 \cdot \frac{Cost_{Base}^2}{\mathcal{C}_{ATTACK}^2}\right) - 172.747 - 40 \cdot \ln\left(\frac{Cost_{Base}}{\mathcal{C}_{ATTACK}}\right)
\end{aligned}
$$
$$(12)$$

Using Equations 11 and 12 we can now produce the optimal model of drones to be deployed, and in what numbers, for any given threat[1]: for example, if the best available drone model costs \$5,000 per unit then for threats of potential damages that are estimated as \$600,000 (which reflect a ratio of $\frac{1}{120}$ between the cost of best unit and the incident's damages) we should use approximately 25 units of the type that costs \$1,100 each (and deploy them according to the selected deployment scheme described above). However, for threats estimated at \$10,000,000 in total damages (i.e. a ratio of $\frac{1}{2000}$) an optimal monitoring system should be comprised of approximately 160 units, of the type that costs \$200 each.

Note that the previous example depends of course on the assumption of quadratic relation between cost and quality of the monitoring units, as well as on the assumption regarding the deployment scheme (i.e. the parameters of the *Gompertz Model*). For any change in any for these assumptions, new corresponding solutions to the optimal monitoring problem can be easily generated using the model.

---

[1]Our model assumes continuous selection of drones' types. In reality of course there is a finite number of drones model. Therefore, after producing the optimal value for the cost of a single drone, we would select the two available models closest in cost to this optimal value (namely, the more expensive cheaper one, and the cheapest more expensive one), assign their cost in the model, and select the one for which the merit function is higher.

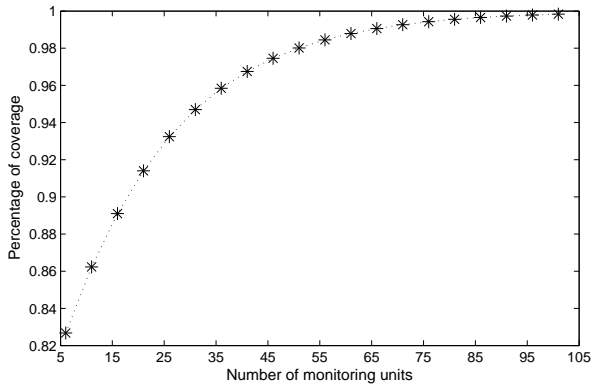

Figure 3: An illustration of the *Gompertz* function $e^{-0.2 \cdot e^{-0.05 \cdot t}}$, used to simulate the increase in monitoring coverage as a function of the increase in number of drones, assumed in the case study of Section 6.

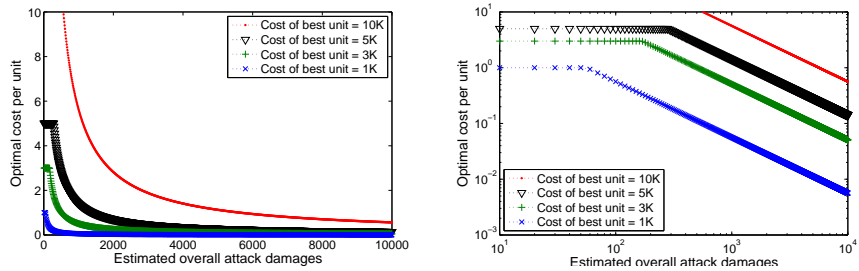

Figure 4: An illustration of Equation 11, denoting the optimal cost of a single drone, as a function of the estimations of the overall damages of a successful attack (from $10K ro $10M). The optimal cost of a unit is shown for 4 possible values of $Cost_{Base}$, the cost of the best unit available. All monetary units are given in $1K. Right chart uses a double log scale.

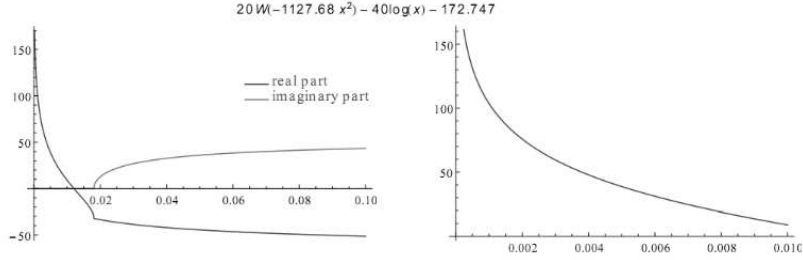

Figure 5: An illustration of Equation 12, showing the optimal number of drones as a function of the ratio between $Cost_{Base}$, the cost of the best drone model available, and the estimation regarding the overall damages of a successful attack (the ratio is denoted by $x$). The left chart contains both the real and imaginary parts of the solution. The right chart shows the segment between 0 and 0.01, where the imaginary part equals zero.

# 7 Case Study II – Real World Transportation Network Monitoring

In the previous sections we have shown an analytic method for optimizing the number and types of monitoring units for any given deployment scheme (defined by the values of Gomperz function's parameters $a$, $b$ and $c$ that set the relation between the amount of monitoring resources and the increase in monitoring efficiency).

In this section we discuss several methods for deploying a given number of monitoring units. The goal in all three methods is the same — maximizing the probability that a target randomly moving in one of the network's edges, is detected.

The most trivial method for deploying $n$ monitoring units is of course, the random deployment method. In other words, positioning the units in a randomly selected nodes in the network. As we will soon see, this method achieves surprisingly well performance in some cases. However, this method can be improved by taking into account certain features of the network, resulting in improved performance in many cases.

In the coming sections we will see several such improvements, all based on traffic-oriented improvement of the standard *Betweenness Centrality* method.

We then show that all methods shown in this section can be efficiently approximated using the *Gompertz* function $y = ae^{be^{ct}}$. Hence, any of those method can be represented by a set of values of $a$, $b$ and $c$, as required by our proposed model. In addition, note that the last deployment method we present here achieves remarkable results (in terms of prediction of the real traffic flow, by the analysis of the network's topology). Therefore, this method can be assumed as a standard deployment scheme for transportation networks, even regardless of the scope of this work.

The data used for the evaluation of this analysis was derived from a large number of location-reports, collected from cell phones in Israel, and is described and discussed in length in Section 8. Parts of the following analysis can also be found in [71] and [18].

## 7.1 Betweenness Centrality vs. Traffic Flow

*Betweenness Centrality* (BC) stands for the ability of an individual node to control the communication flow in the networks and is defined as the total fraction of shortest paths between each pair of vertices that pass through a given vertex [20, 44]. In recent years Betweenness was extensively applied for the analysis of various complex networks [23, 81] including among others social networks [77, 87], computer communication networks [42, 93], and protein interaction networks [28]. Holme [51] have shown that Betweenness is highly correlated with congestion in particle hopping systems. Extensions of the original definition of BC are applicable for directed and weighted networks [29, 89] as well as for multilayer networks where the underlying infrastructure and the origin-destination overlay are explicitly defined [70].

Let $G = (V, E)$ be a directed transportation network where $V$ is the set of junctions and $E$ is the set of directed links as described in Section 8. Let $\sigma_{s,t}$ be the number of shortest paths between the origin vertex $s \in V$ and the destination vertex $t \in V$ (in some applications the shortest path constraint can be relieved to allow some deviations from the minimal distance between the two vertices). In the rest of this paper we will refer to the shortest or "almost" shortest paths between two vertices as *routes*. Let $\sigma_{s,t}(v)$ be the number of routes from $s$ to $t$ that pass through the vertex $v$. The Betweenness Centrality can hence be expressed by the following equation:

$$BC(v) = \sum_{s,t \in V} \frac{\sigma_{s,t}(v)}{\sigma_{s,t}}. \tag{13}$$

Note that in this definition we include the end vertices ($s$ and $t$) in the computation of Betweenness since we assume that vehicles can be inspected also at their origin and at the point of their destination.

After computing the Betweenness Centrality for the given transportation network, we can easily see that the distribution of Betweenness Centrality follows a power law (Figure 6). Long tail distributions such as the power law suggest that there is a non negligible probability for existence of vertices with very high Betweenness Centrality. This is in contrast to the exponential flow distribution depicted in Figure 19. The different nature of these two distributions suggests that BC as defined above will overestimate the actual traffic flow through nodes especially for the most central vertices.

Next we would like to check the correlation between BC and traffic flow. Although the correlation is significant the square error is very low ($R^2 = 0.2021$) as shown in Figure 7 (a). Every point in this Figure represents a vertex with the x-axis corresponding to the measured traffic flow and y-axis corresponding to the computed BC.

We now discuss augmented variants of the Betweenness Centrality measure that significantly improve the correlation with the traffic flow.

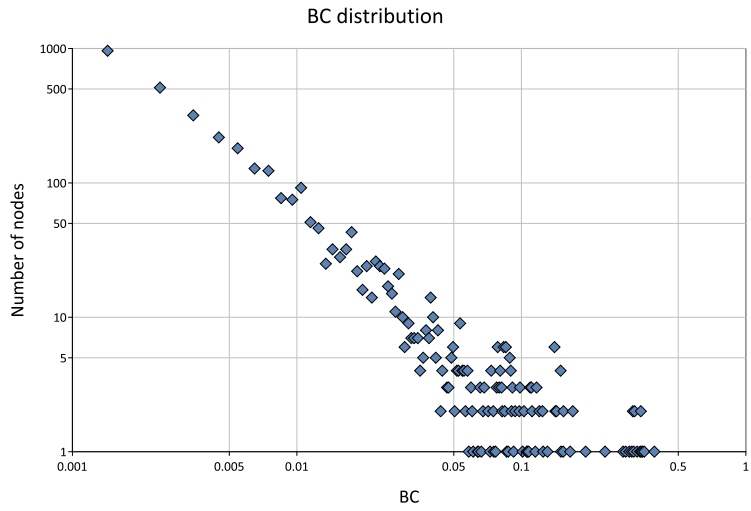

Figure 6: Power law distribution of Betweenness Centrality

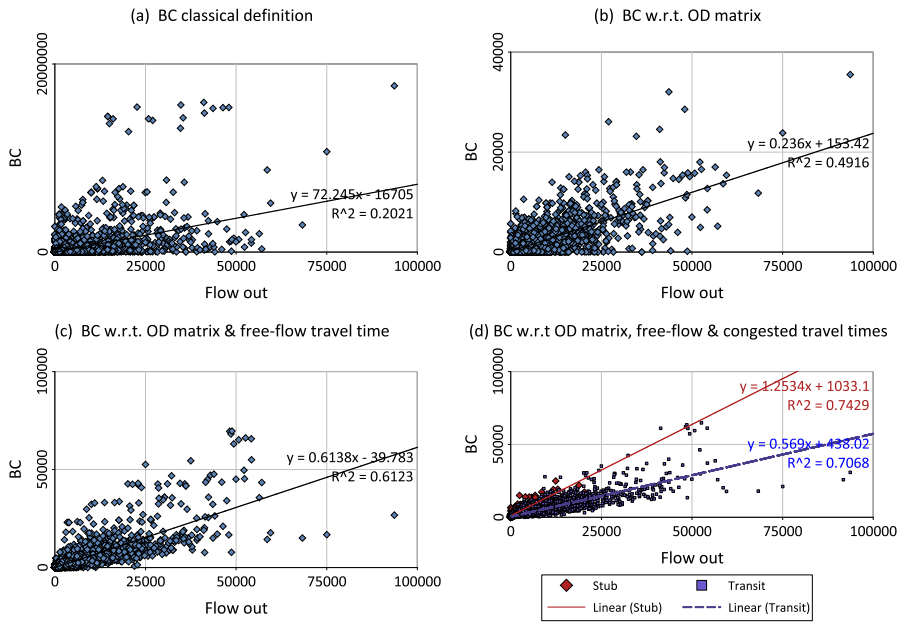

Figure 7: Correlation of flow through nodes and Betweenness Centrality

## 7.2  *Origin-Destination* based Betweenness Centrality

BC definition according to Equation 13 BC assumes equal weights of routes between every pair of vertices in the network. In other words every vertex acts as an origin and as a destination of traffic. We would like to utilize the measured origin-destination (OD) flow matrix in order to prioritize network regions by their actual use. For this, we shall use the following altered definition for Betweenness, as suggested in [70]:

$$BC(v) = \sum_{s,t \in V} \frac{\sigma_{s,t}(v)}{\sigma_{s,t}} \cdot OD_{s,t} \qquad (14)$$

where $OD$ is the actual measured origin-destination matrix. This method produces a better correlation ($R^2 = 0.4916$) between the theoretic (BC) and the measured traffic flow (see Figure 7 (b)).

## 7.3  Improving Betweenness Centrality using Travel Properties

**Shortest Routes based on Time to Travel**  In order to further improve our ability to estimate the predicted network flow using the network's topology, we note that both BC calculation methods (Equations 13 and 14 above) assume that routes are chosen according to shortest path strategy based on hop counting. In this section, we retain the shortest path assumption but use weighted links for calculating the Betweenness score. One option is to use the length of the road segments as their weights for the shortest path calculations. Shortest path algorithms (such as Dijkstra's or Bellman-Ford's) are able to consider only one distance weight on links when computing the shortest path to a destination. We shall therefore assume that the primary heuristic guiding people when they chose a route is the time required to reach their destination, and recompute the BC on the directed transportation, weighting links by their free-flow travel time.

Let $BC^{ft}(v)$ denote the Betweenness of a node $v$ computed w.r.t. the free-flow travel time. Figure 7 (c) shows significant improvements in the correlation between the measured traffic flow and the theoretical $BC^{ft}$ values computed w.r.t the OD matrix and free-flow travel time link weights ($R^2 = 0.6123$). We can see that there are few nodes whose flow was significantly underestimated by the BC measure. Notice that there are also several nodes whose flow was actually overestimated. This can be explained by the fact that people do not travel strictly via shortest paths, but may have various deviations. In particular the deviations form shortest paths are affected by the day time and the day of week.

**Peak-Hours Aware** **Betweenness Centrality**  It is a reasonable assumption that during peak hours travelers will choose to avoid the congested roads and choose their routes based on the congested travel time rather than on the free-flow travel times. Let $BC^{ct}(v)$ denote the Betweenness of a node $v$ computed w.r.t. the congested time. Computing Betweenness using only the congested travel time weights results in $R^2 = 0.7096$. Although peak hours are relatively small fraction of the day, most vehicles travel at these hours. This is the reason for higher correlation of $BC^{ct}$ with the measured traffic flow.

We shall now combine both the Betweenness Centrality computed w.r.t. the free-flow travel time and the congested time by taking a weighted average, namely :

$$BC(v) = \alpha \cdot BC^{ft}(v) + (1 - \alpha) \cdot BC^{ct}(v)$$

where $\alpha$ denotes the relative fraction of vehicles traveling during the free-flow periods. The resulting centrality index can achieve higher correlation with the measured average traffic flow. The maximal correlation of $R^2 = 0.7285$ is obtained for $\alpha = 0.25$ as shown in the Figure 8.

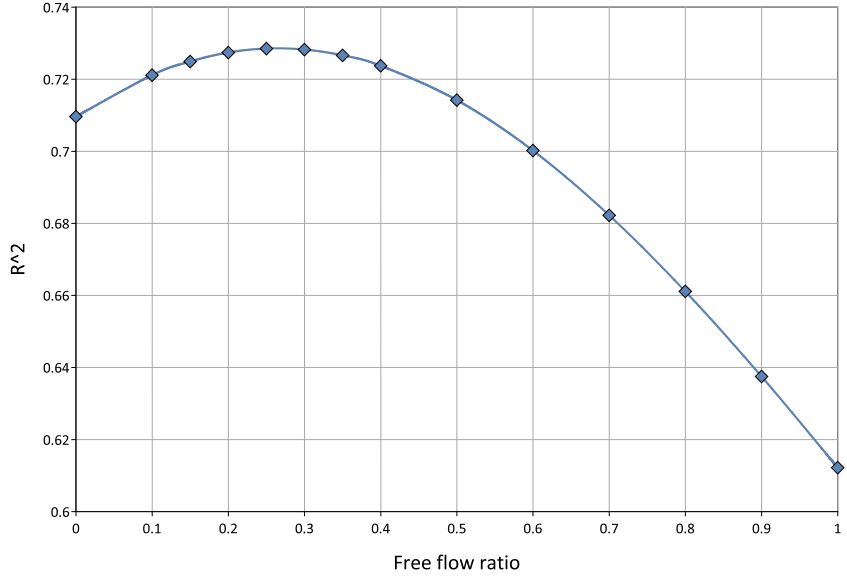

Figure 8: Squared error ($R^2$) as the function of the free flow traffic fraction ($\alpha$).

**Separating *Stubs Nodes* from *Transit Nodes*** Nodes in the dataset were divided into two groups : *stub nodes* and *transit nodes*. A Stub node (i.e., centroid) is a node that is an origin or a destination of the traffic (as seen in the Origin-Destination matrix). These nodes account for approximately 10% of the network's nodes. All other nodes (namely, nodes that generate insignificant or no outgoing or incoming routes) are called Transit nodes, as they only forward traffic and do not generate or consume it.

Figure 7 (d) presents the correlation that is received when the two groups of nodes are being processed separately. Specifically, the results show a $R^2 = 0.7068$ for the Transit nodes and a $R^2 = 0.7429$ for the Stub nodes.

***Mobility Oriented* Betweenness Centrality** As each type of roads has a different functional class, we shall further improve our flow prediction by examining the Betweenness values achieved when calculating it for every group separately. The results of the correlation that is achieved using this method are presented in Figure 9. We can clearly see that for the more important roads (namely, those with lower type number, representing a more infrastructural role in the transportation network) this technique yields $R^2$ values that are consistently above 0.74, reaching 0.83(!) for road of types 2 and 9 (note that roads of type 90 are fictive roads with infinite capacity that were artificially added in order to connect distinct regions in the network).

It should be noted that each node may have incoming roads of different types. Each plot corresponds to a set of nodes whose max incoming road type is as specified. In addition, the BC calculations were not made for each set of nodes separately — BC was computed for the complete network, while the correlations were computed separately for each type.

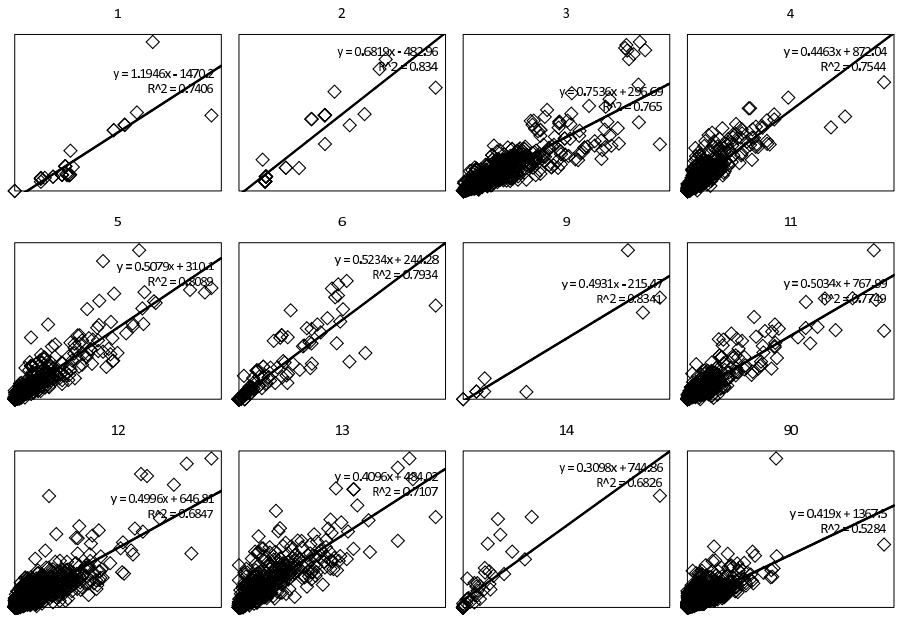

Figure 9: Correlation of flow through nodes and Betweenness computed separately for different *types* of links.

## 7.4 Optimizing the Locations of Surveillance and Monitoring Stations

In this section we use the Group variant of shortest path Betweenness Centrality (GBC) [41] as an estimate for the utility of collaborative monitoring. In other words, we are interested in verifying that given some mobile agent that we are interested in intercepting, we position the monitoring stations in a way that maximize the chance the agent would be captured, given the traffic patterns of the transportation network. In this case, however, significant computational complexity issues arise, rendering the generation of an optimal solution impractical in real time by conventional tools that are based mostly on behavioral based modeling. Using Group Between Centrality we propose a way to generate efficient approximations of the optimal solution to this optimization problem.

GBC of a given group ($M \subseteq V$) of vertices accounts for all routes that pass through *at least one* member of the group. Let $\sigma_{s,t}$ and $\sigma_{s,t}(M)$ be the number of routes from $s$ to $t$ and the number of routes from $s$ to $t$ passing through at least one vertex in $M$ respectively :

$$GBC(M) = \sum_{s,t \in V} \frac{\sigma_{s,t}}{\sigma_{s,t}(M)} \cdot OD_{s,t} \tag{15}$$

GBC can be efficiently computed using the algorithm presented in [68].

Assuming the routes are weighted by the origin destination flow in transportation networks, GBC will account for the net number of vehicles that are expected to pass by the monitors during an hour. This net number is different from the total number of vehicle passing by the monitors since the same vehicle can pass by several monitors during a single trip. For example, searching for a suspected escaping terrorist car, one would like to avoid stopping the same vehicle twice and increase the number of distinct vehicles that were inspected. It is therefore important to maximize the GBC value of the set of inspection stations given the number of stations deployed.

Several combinatorial optimization techniques can be used to find a group of nodes of given size that has the largest GBC. In the following discussion we refer to a greedy approximation algorithm for the monitors location optimization problem (Greedy) [37], a classical *Depth First Branch and Bound* (DFBnB) heuristic search algorithm [58], and recently proposed *Potential Search* [80].

The *Greedy* approximation algorithm chooses at every stage the node that has the maximal contribution to the GBC of the already chosen group. The approximation factor of the *Greedy* algorithm as reported in [37] is $e - \frac{1}{e}$.

Both the heuristic search algorithms *DFBnB* and the *Potential Search* provably find the group having the maximal GBC. The *Greedy* algorithm and *DFBnB* were previously compared in [69] in the context of monitoring optimization in computer communication networks. Given the fact that finding a group of a given size having the maximal GBC is a hard problem, the greedy algorithm is good enough for any practical purpose (the hardness of the problem can be proven by a straightforward reduction from the Minimal Vertex Cover problem that the problem of maximizing GBC is NP-Complete). Figure 10 presents the results of selecting one to 39 inspection locations using the greedy algorithm.

In certain cases (such as in various homeland security applications) deployment of monitoring systems are often done under tight timing conditions, as a result of new intelligence information. Therefore, any optimization method should provide close-to-real-time capabilities. In this context, it is interesting to note that both the *DFBnB* and the *Potential* algorithms are anytime search algorithms [94]. Their execution can be stopped at any point of time, yielding the best solution found so far. Therefore, in the following experiments we limit the search time to one hour, simulating a quasi-real-time optimization constraint. Still, as can be seen in figure 11 the running time of then *Greedy* algorithm is by far lower than one hour, for the entire Israeli transportation system.

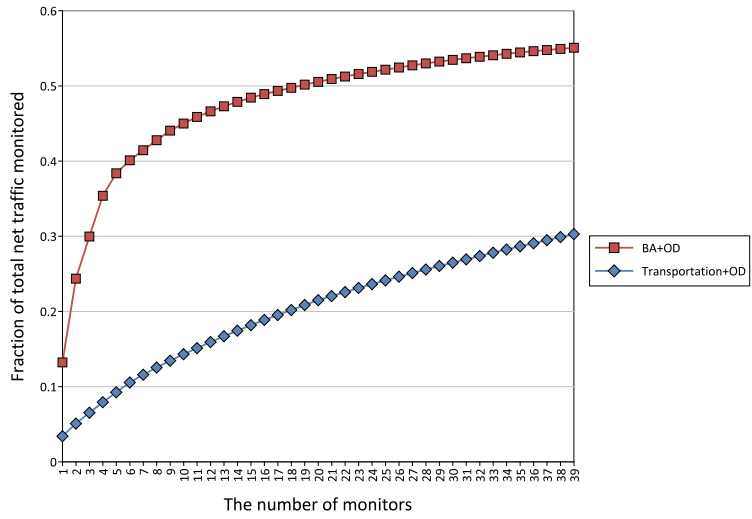

Figure 10: The total net traffic flow that passes by monitors as a functions of the number of monitors. As expected the marginal value of additional monitors gradually decreases as more of them are added reaching potential traffic coverage of 30% when 39 monitoring stations are deployed.

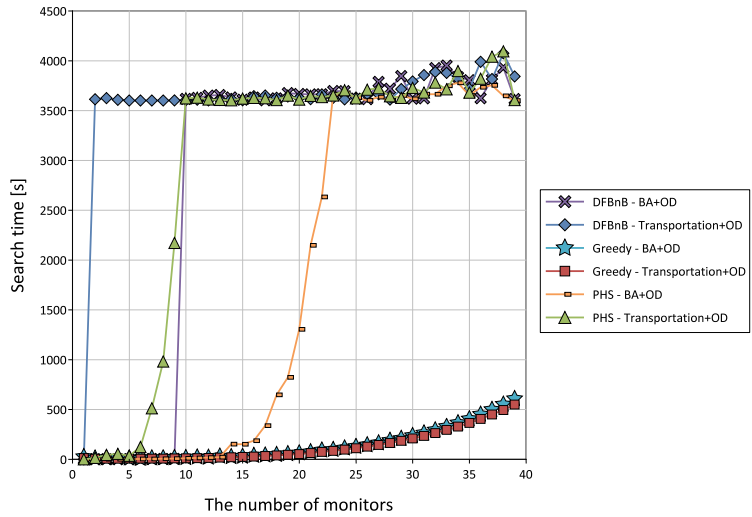

Figure 11: The time (in seconds) that the search algorithms were executed as a function of the number of monitors. Note that due to the high complexity nature of the search algorithms, they present a phase transition around a tipping point (specific to each algorithm) in the number of monitors to be optimized.

When *DFBnB* and *Potential Search* algorithms cannot complete the search process within the given time bounds they produce a close to optimal solution and an estimate of its optimality (i.e. certificate). The certificate is computed by dividing the best solution found so far by the upper bound on the optimal solution. The upper bound is computed using admissible heuristic functions and is maintained by the search algorithms for efficient pruning the search space. Figure 12 shows that *Potential Search* produces higher certificates for its solutions within the one hour time bound for all sizes of the monitors deployment.

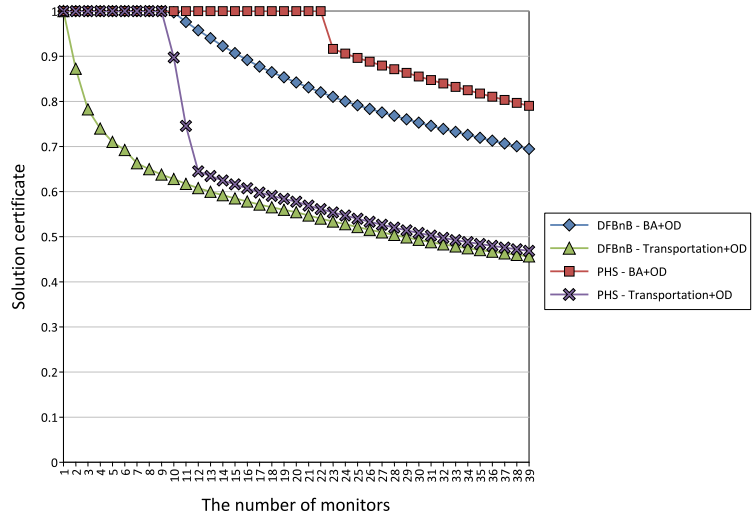

Figure 12: The minimal quality of the solution (fraction of the upper bound) as a function of the number of monitors.

## 7.5 Modeling Various Deployment Schemes as a *Gompertz* Function

Figure 13 demonstrates the performance of our monitoring method, by showing the percentage of traffic monitored as a function of the number of monitors, for several deployment schemes: (a) Group Betweenness, (b) Betweenness, and (c) Random deployment. The benefits of the proposed method can clearly be seen from this chart.

Remarkably, using GBC based deployment strategy it is possible to cover the vast majority of the traffic in the analyzed network (detecting any threat in probability of approximately 0.7) using only 70 monitors. This result is most probably due to the relatively small number of origins and destinations in the analyzed network. 680 origins/destinations account for a little bit more than 10% of the network nodes.

BC based strategy produces relatively high quality deployments for small number of monitors (less than five). However, when 10 or more monitors need to be located random deployment is on average as effective as choosing the most central intersections.

Moreover, for large numbers of monitors (more than 70-80) random deployment, although the simplest strategy, achieves coverage results that are very similar to choosing the most central intersections. This result may seem surprising but in fact it is absolutely reasonable. Central intersections tend to lay on the arterial roads and usually are quite close to each other. This results in reduced marginal utility of each additional junction joining the deployment.

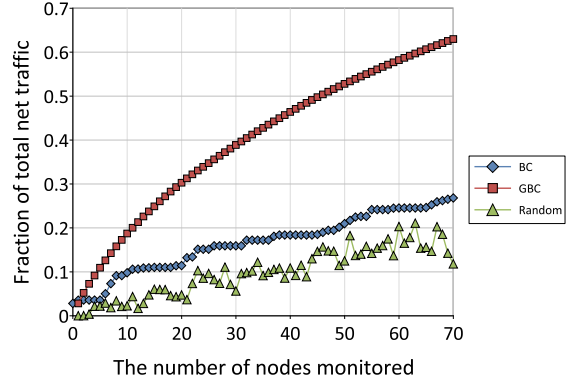

Figure 13: The figure presents the results of deployment optimization performed on the Israeli transportation network with average travel times computed using state of the art traffic assignment model. Flows and the utility of the deployment were estimated using *Betweenness Centrality* and *Group Betweenness Centrality* models, and compared also to the random deployment model. Whereas the *BC* algorithm had chosen the locations for monitoring units according to the most central intersection based on their BC values, the *GBC* deployment was a greedy algorithm that tried to maximize the net-number of vehicles passing by the monitors. The benefits of the *GBC* strategy is clearly shown, as well as the ability to extrapolate this correlation between number of monitoring units and monitored traffic percentage, in order to find the minimal number of monitoring units required in order to guarantee certain levels of coverage.

We demonstrate Equation 3 using the results presented in Figure 13. For this, we need to calculate the regression of the simulated measurements presented in Figure 13 to the *Gompertz* function. This is presented in Figure 14. The regression yields the following results :

$$
\begin{aligned}
\textbf{Random deployment : } M(n) &= 0.24e^{-2.73e^{-0.03n}} \\
\textbf{BC deployment : } M(n) &= 0.51e^{-2.14e^{-0.02n}} \\
\textbf{GBC deployment : } M(n) &= 0.89e^{-2e^{-0.04n}}
\end{aligned}
\tag{16}
$$

At this point, we can assign the values of $a$,$b$ and $c$ in Equation 3 and get the optimal number of monitoring units, for any ratio between the cost of a single monitoring unit and the expected cost of a successful attack. This is presented in 15, where the benefit of the GBC deployment scheme can clearly be seen, as it enables the use of much more expensive monitoring units (which make little sense in using, when the cost of using them is greater than the potential damage of a missed detection — due to low monitoring efficiency).

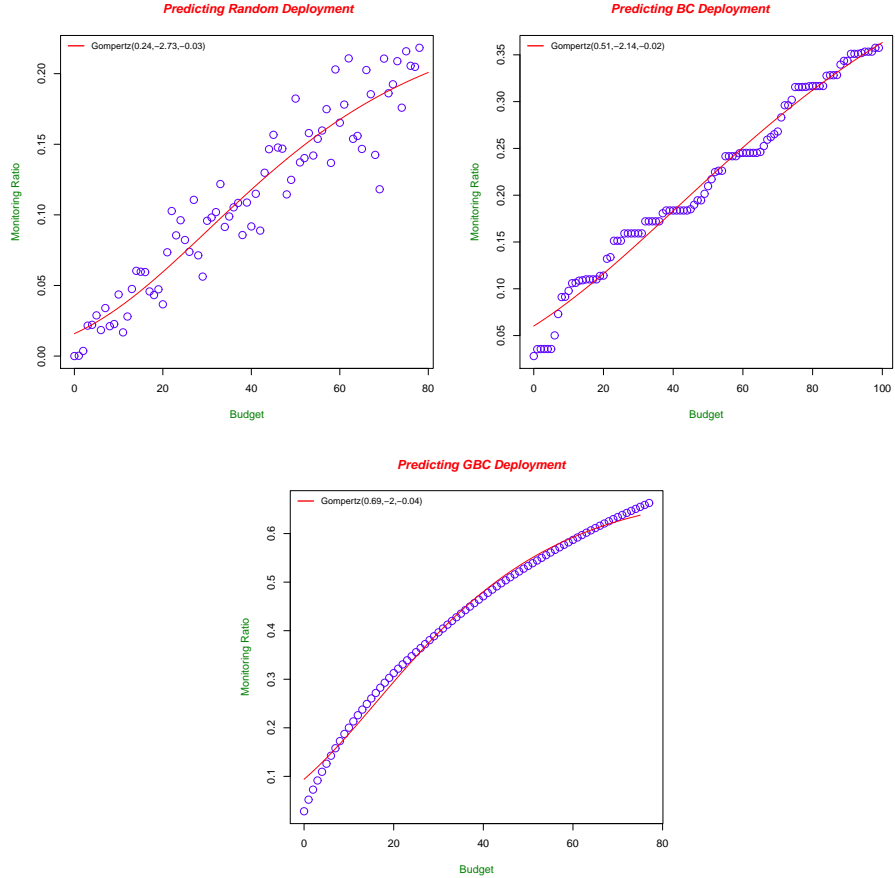

Figure 14: *Gompertz* regression of the measurements presented in Figure 13 — Random deployment, BC deployment and GBC deployment.

# 8   Transportation Network Dataset

In this section we evaluate our proposed model using a real-world transportation dataset, containing data regarding the Israeli roads structure, as well as the traffic through them (simulating the potential targets that need to be detected or monitored). This section presents the dataset and its various aspects, and discusses in great length its network properties – from which the optimal deployment of the drones swarm can be derived.

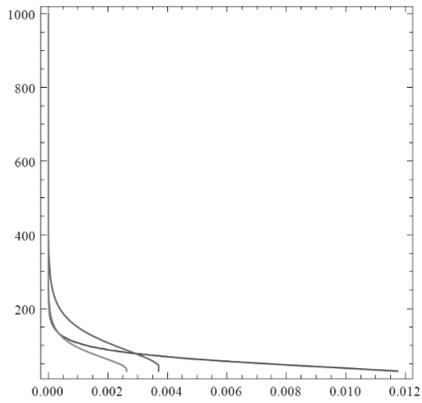

Figure 15: The optimal number of monitoring units as a function of the ratio between the cost of a single monitoring unit, and the cost of a successful attack, for the *Gompertz* regressions presented in Figure 14 for random deployment (green), BC deployment (purple), and GBC deployment (blue). Note how both the random deployment and the BC deployment schemes can be used only for very low-cost monitoring units, whereas the GBC deployment scheme enables the use of much more expensive monitoring units (due to its increased efficiency in guaranteeing high monitoring coverage).

The widespread use of cellular phones in Israel enables the collection of accurate transportation data. Given the small size of the country, all cellular companies provide national wide coverage. As shown in [25], the penetration of cellular phones to the Israeli market is very high, even to lower income households, and specially among individuals in the ages of 10 to 70 (the main focus of travel behavior studies). Such penetration enables a comprehensive study of travel behavior that is based on the mobility patterns of randomly selected mobile phones in the Israeli transportation system. This data was shown in [25] and [92] to provide a high quality coverage of the network, tracking $94\%$ of the trips (defined as at least 2km in urban areas, and at least 10km in rural areas). The resulting data contained a wealth of traffic properties for a network of over 6,000 nodes, and 15,000 directed links. In addition, the network was accompanied with an Origin Destination (OD) matrix, specifying start and end points of trips.

The network was created for the National Israeli Transportation Planning Model. In urban areas the network contains arterial streets that connect the interurban roads. For each link of the network, there is information about the length (km), hierarchical type, free-flow travel time (min), capacity (vehicles per hour), toll (min), hourly flow (vehicles per hour), and congested travel time (min). The hourly flows and congested travel times were obtained from a traffic assignment model that loads the OD matrix on the network links.

## 8.1 Network Structure

Based on the dataset described above we have created a network structure, assigning running indices from 1 to 6716 to the nodes (junctions). We have examined the directed variant of the network where each road segment between two junctions was represented as either one or two directed links between the respective nodes.

In order to get a basic understanding of the network we first extracted and studied several of its structural properties (see Table 1). We have partitioned the network into structural equivalence classes of the nodes and bi-connected components and computed the Betweenness Centrality indices of the nodes (Betweenness Centrality is an important network feature that measures the portion of shortests paths between all the pairs of nodes in the network, that pass through a particular node. See more details in [44, 60, 62]). Structurally equivalent vertices have exactly the same neighbors and the set of these vertices is called a structural equivalence class. As can be seen from Table 1 the number of structural equivalence classes is roughly the number of vertices in the network and the size of the largest class is three. This means that there are no "star-like" structures in the network and alternative paths between any two vertices are either longer than two hops or have other links emanating from the intermediate vertices. On the other hand the number of biconnected components in the network is low compared to the number of nodes, meaning that there are significant regions of the network that can be cut out by merely disconnecting a single node.

## 8.2 Congestions

In this paper we define the impact of congestion as the difference between the time to travel through a congested link and the free-flow time to travel. Congestion of a junction can be either inbound or outbound. Inbound congestion is the sum of all congestions on inbound links of some junction. Figure 17 presents the distribution of congestion on network nodes (junctions). Power law nature of this distribution means that vast majority of nodes are not congested but there are a few nodes whose congestion can be arbitrarily large. Based on the *Wardrop's User Equilibrium* [45] this also implies a low number of yet significant deviations between the routes chosen by travelers during free-flow and during congestions. In Section 7.3 we use this fact to merge between two routing strategies.

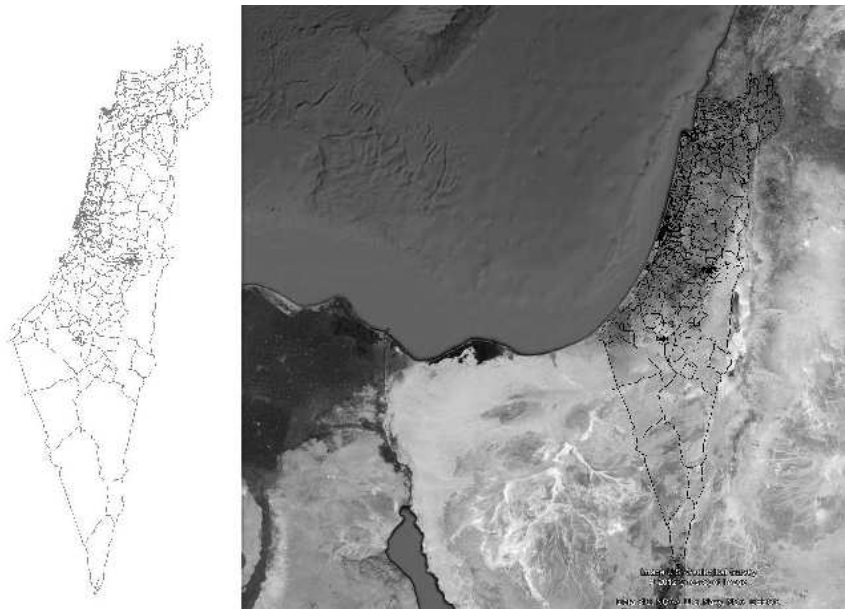

Figure 16: A map of the Israeli transportation network that was used for this paper.

Table 1: Structural properties (Israeli transportation network).

| | |
|---|---|
| Nodes | 6716 |
| Edges (undirected representation) | 8374 |
| Edges (directed representation) | 15823 |
| Number of structural equivalence classes | 6655 |
| Largest equivalence class | 3 |
| Number of bi-connected components (BCC) | 931 |
| Avg BCC size | 8.2 |
| Largest BCC | 5778 |

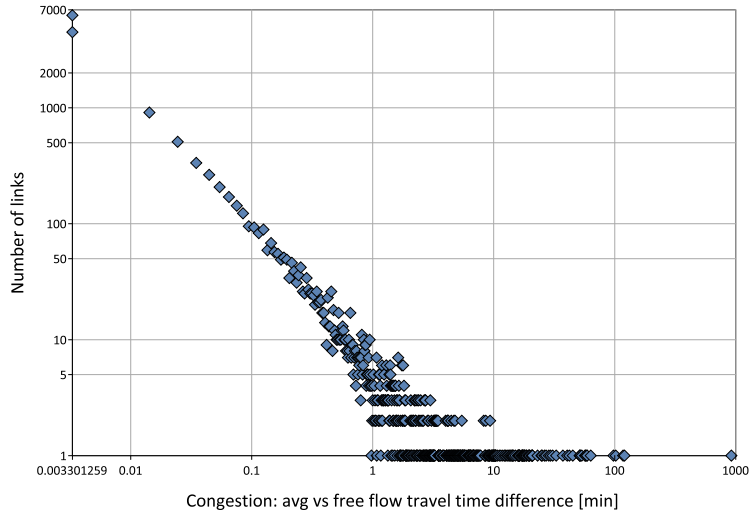

Figure 17: Power law distribution of congestion.

## 8.3   Flow

The analyzed dataset contains traffic flow through links provided as the number of vehicles per hour. We compute the total inbound flow through a node by summing flows on all of its inbound links, where outbound flow is computed symmetrically.

Unless a specific junction is a source or a destination of traffic we expect the inbound flow to be equal to the outbound flow. Figure 18 demonstrates the correlation between inbound and outbound flow. We see that vast majority of the nodes are located on the main diagonal, however, there are some deviations, caused by the fact that the data represents average measurements that were carried out along a substantial period of time.

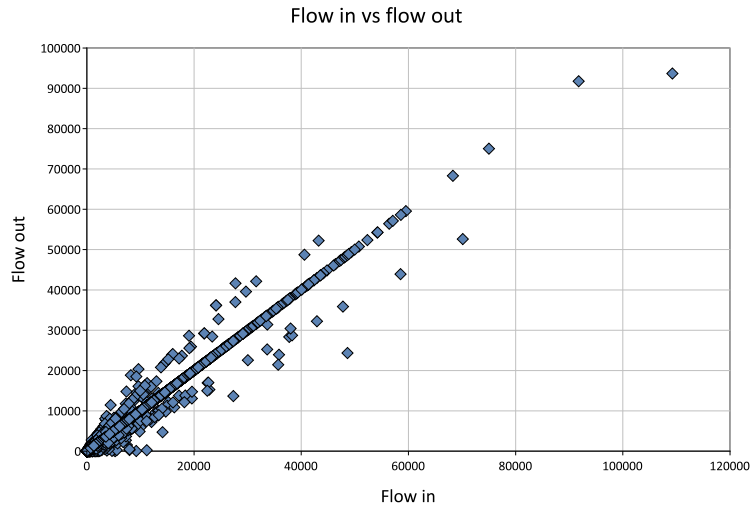

Figure 18: Incoming vs. outgoing flow for each node.

Figure 19 presents the distribution of inbound flow on network nodes. This distribution is exponential, meaning that a vast majority of nodes have little flow through them. However, in contrast to network congestion, there are no "unbounded fluctuations", i.e. the flow through the most "busy" junctions is not as high as can be expected from the power law distribution of betweenness and congestions (Figures 17 and 6). In fact, congestions significantly limit the flow through the busiest junctions, which subsequently is the reason we do not see the long tail in flow distribution.

# 9 Conclusions

In this paper we have discussed the problem of optimizing the type, number and locations of surveillance drones trying to detect maneuvering targets that move on top of a pre-defined transportation network. We have presented a model for analytically generating an optimal monitoring strategy, flexible enough to support various models of drones (of different costs and performance specifications), as well as any deployment scheme used. For each threat, the model produces an optimal strategy, based on the estimation of the threat's overall potential damages.

We have validated our model using a variety of deployment schemes, among which was an extremely efficient scheme we have developed, which is a transportation oriented variant of the Network Betweenness measure. We have then evaluated the model, as well as this deployment scheme, using a comprehensive dataset that covers the Israeli transportation network, and demonstrated how the optimal locations of any (reasonable) amount of drones can be approximated in high accuracy. The method proposed in this paper can now be used in order to generate highly efficient dynamic monitoring strategies, for a multitude of real-time threats, of different operational parameters.

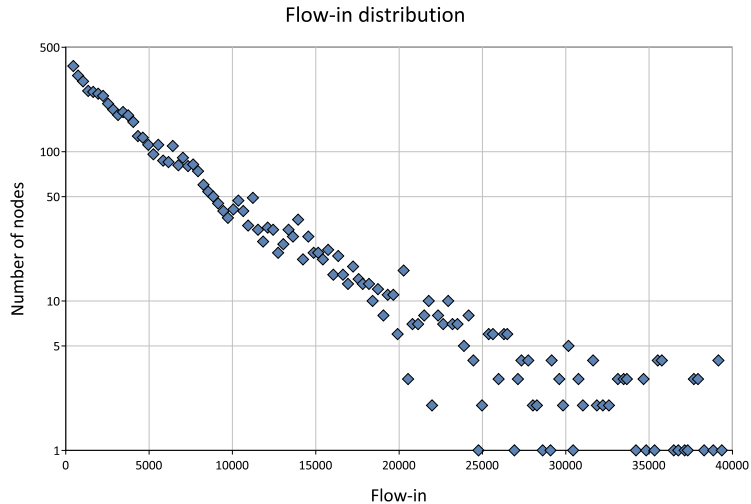

Figure 19: Exponential distribution of traffic flow through nodes.

A more theoretical approach to this problem that studies the complexity of all possible strategies for a collaborative monitoring of a given area at which an unknown number of targets dynamically maneuver can be found in [16]. An additional similar variant to this problem is the search for pollutant emitting vehicles, where the merit function is derived from environments considerations [71]. It is interesting to mentioned that in those variants as well, the topological properties of the network along which the "targets" can move significantly influence the ability of drones to track them, as was pointed out in [8, 9].

In future works we plan to conduct a thorough study of the way this method can be used for resource allocation scenarios involving several types threats, using heterogenous deployment strategies that would utilize drones of different characteristics simultaneously. In addition, we intend to conduct a further analysis of the resilience of the method, under uncertainty (regarding the topology, potential threats, and noises in the deployment scheme).

Finally, it is interesting to note that the problem of finding optimal (and optionally dynamic) monitoring strategy is related to other kinds of monitoring problems, such as "reverse-monitoring" – maximizing the exposure of a certain signal, or campaign to a population [19], or monitoring for evading land targets by a flock of Unmanned

Air Vehicles (UAV). In this problem, however, the fact that the paths of the UAVs is unconstrained (as they are flying in the air) makes the calculation of a near-optimal monitoring strategy fairly easy [12, 13].

# References

[1] E.U. Acar, Y. Zhang, H. Choset, M. Schervish, A.G. Costa, R. Melamud, D.C. Lean, and A. Gravelin. Path planning for robotic demining and development of a test platform. In *International Conference on Field and Service Robotics*, pages 161–168, 2001.

[2] Matthew M Aid. All glory is fleeting: Sigint and the fight against international terrorism. *Intelligence and National Security*, 18(4):72–120, 2003.

[3] S. Alpern and S. Gal. *The Theory of Search Games and Rendezvous*. Kluwer Academic Publishers, 2003.

[4] Y. Altshuler, N. Aharony, M. Fire, Y. Elovici, and A Pentland. Incremental learning with accuracy prediction of social and individual properties from mobile-phone data. *CoRR*, 2011.

[5] Y. Altshuler, N. Aharony, A. Pentland, Y. Elovici, and M. Cebrian. Stealing reality: When criminals become data scientists (or vice versa). *Intelligent Systems, IEEE*, 26(6):22–30, nov.-dec. 2011.

[6] Y. Altshuler, A.M. Bruckstein, and I.A. Wagner. Swarm robotics for a dynamic cleaning problem. In *IEEE Swarm Intelligence Symposium*, pages 209–216, 2005.

[7] Y. Altshuler, M. Fire, N. Aharony, Y. Elovici, and A Pentland. How many makes a crowd? on the correlation between groups' size and the accuracy of modeling. In *International Conference on Social Computing, Behavioral-Cultural Modeling and Prediction*, pages 43–52. Springer, 2012.

[8] Y. Altshuler, I.A. Wagner, and A.M. Bruckstein. Shape factor's effect on a dynamic cleaners swarm. In *Third International Conference on Informatics in Control, Automation and Robotics (ICINCO), the Second International Workshop on Multi-Agent Robotic Systems (MARS)*, pages 13–21, 2006.

[9] Y. Altshuler, I.A. Wagner, and A.M. Bruckstein. On swarm optimality in dynamic and symmetric environments. volume 7, page 11, 2008.

[10] Y. Altshuler, I.A. Wagner, and A.M. Bruckstein. Collaborative exploration in grid domains. In *Sixth International Conference on Informatics in Control, Automation and Robotics (ICINCO)*, 2009.

[11] Y. Altshuler, I.A. Wagner, V. Yanovski, and A.M. Bruckstein. Multi-agent cooperative cleaning of expanding domains. *International Journal of Robotics Research*, 30:1037–1071, 2010.

[12] Y. Altshuler, V. Yanovski, I.A. Wagner, and A.M. Bruckstein. The cooperative hunters - efficient cooperative search for smart targets using uav swarms. In *Second International Conference on Informatics in Control, Automation and Robotics (ICINCO), the First International Workshop on Multi-Agent Robotic Systems (MARS)*, pages 165–170, 2005.

[13] Y. Altshuler, V. Yanovsky, A.M. Bruckstein, and I.A. Wagner. Efficient cooperative search of smart targets using uav swarms. *ROBOTICA*, 26:551–557, 2008.

[14] Y. Altshuler, V. Yanovsky, I. Wagner, and A. Bruckstein. Swarm intelligence-searchers, cleaners and hunters. *Swarm Intelligent Systems*, pages 93–132, 2006.

[15] Yaniv Altshuler and Alfred M. Bruckstein. The complexity of grid coverage by swarm robotics. In *ANTS 2010*, pages 536–543. LNCS, 2010.

[16] Yaniv Altshuler and Alfred M. Bruckstein. Static and expanding grid coverage with ant robots: Complexity results. *Theoretical Computer Science*, 412(35):4661–4674, 2011.

[17] Yaniv Altshuler, Michael Fire, Nadav Aharony, Zeev Volkovich, Yuval Elovici, and Alex Sandy Pentland. Trade-offs in social and behavioral modeling in mobile networks. In *Social Computing, Behavioral-Cultural Modeling and Prediction*, pages 412–423. Springer, 2013.

[18] Yaniv Altshuler, Rami Puzis, Yuval Elovici, Shlomo Bekhor, and Alex Sandy Pentland. On the rationality and optimality of transportation networks defense. *Securing Transportation Systems*, pages 35–63.

[19] Yaniv Altshuler, Erez Shmueli, Guy Zyskind, Oren Lederman, Nuria Oliver, and Alex Pentland. Campaign optimization through behavioral modeling and mobile network analysis. *Computational Social Systems, IEEE Transactions on*, 1(2):121–134, 2014.

[20] J. M. Anthonisse. The rush in a directed graph. Technical Report BN 9/71, Stichting Mathematisch Centrum, Amsterdam, 1971.

[21] R.C. Arkin. Integrating behavioral, perceptual, and world knowledge in reactive navigation. *Robotics and Autonomous Systems*, 6:105–122, 1990.

[22] Desmond Ball et al. *Burma's Military Secrets: Signals Intelligence (SIGINT) from 1941 to Cyber Warfare*. White Lotus Press, 1998.

[23] M. Barthélemy. Betweenness centrality in large complex networks. *The European Physical Journal B – Condensed Matter*, 38(2):163–168, March 2004.

[24] R. Bejar, B. Krishnamachari, C. Gomes, and B. Selman. Distributed constraint satisfaction in a wireless sensor tracking system. In *Proceedings of the IJCAI-01 Workshop on Distributed Constraint Reasoning*, 2001.

[25] Shlomo Bekhor, Yehoshua Cohen, and Charles Solomon. Evaluating long-distance travel patterns in israel by tracking cellular phone positions. *Journal of Advanced Transportation*, pages n/a–n/a, 2011.

[26] Donald J. Berndt, Alan R. Hevner, and James Studnicki. Bioterrorism surveillance with real-time data warehousing. In *Proceedings of the 1st NSF/NIJ conference on Intelligence and security informatics*, ISI'03, pages 322–335, Berlin, Heidelberg, 2003. Springer-Verlag.

[27] Confessore G. Reverberi P. Bianco, L. A network based model for traffic sensor location with implications on o/d matrix estimates. *Transportation Science*, 35(1):50–60, 2001.

[28] P. Bork, L. J. Jensen, C. von Mering, A. K. Ramani, I. Lee, and E. M. Marcotte. Protein interaction networks from yeast to human. *Curr. Opin. Struct. Biol.*, 14(3):292–299, 2004.

[29] U. Brandes. On variants of shortest-path betweenness centrality and their generic computation. *Social Networks*, 30(2):136–145, 2008.

[30] Z. Butler, A. Rizzi, and R. Hollis. Distributed coverage of rectilinear environments. In *Proceedings of the Workshop on the Algorithmic Foundations of Robotics*, 2001.

[31] Daniel Byman. Why drones work. *Foreign Affairs*, 92(4):32–43, 2013.

[32] H. Chen, Fei-Yue Wang, and D. Zeng. Intelligence and security informatics for homeland security: information, communication, and transportation. *Intelligent Transportation Systems, IEEE Transactions on*, 5(4):329 – 341, dec. 2004.

[33] Robert Chesney. Military-intelligence convergence and the law of the title 10/title 50 debate. *Journal of National Security Law and Policy*, 5:539, 2012.

[34] D. Chevallier and S. Payandeh. On kinematic geometry of multi-agent manipulating system based on the contact force information. In *The Sixth International Conference on Intelligent Autonomous Systems (IAS-6)*, pages 188–195, 2000.

[35] Lt Kendra LB Cook. The silent force multiplier: the history and role of uavs in warfare. In *Aerospace Conference, 2007 IEEE*, pages 1–7. IEEE, 2007.

[36] L. Damianos, J. Ponte, S. Wohlever, F. Reeder, D. Day, G. Wilson, and L. Hirschman. Mitap for bio-security: A case study. *AI Magazine*, 23(4):13–29, 2002.

[37] S. Dolev, Y. Elovici, R. Puzis, and P. Zilberman. Incremental deployment of network monitors based on group betweenness centrality. *Inf. Proc. Letters*, 109:1172–1176, 2009.

[38] A. d'Onofrio. A general framework for modeling tumor-immune system competition and immunotherapy: Mathematical analysis and biomedical inferences. *Physica D*, 208:220–235, 2005.

[39] Bell M.G. Grosso S. Ehlert, A. The optimisation of traffic count locations in road networks. *Transportation Research Part B*, 40(6):460–479, 2006.

[40] G.M. Erickson, P.J. Currie, B.D. Inouye, and A.A. Winn. Tyrannosaur life tables: An example of nonavian dinosaur population biology. *Science*, 313(5784):213–217, 2006.

[41] M. G. Everett and S. P. Borgatti. The centrality of groups and classes. *Mathematical Sociology*, 23(3):181–201, 1999.

[42] M. Faloutsos, P. Faloutsos, and C. Faloutsos. On power-law relationships of the internet topology. *SIGCOMM Comput. Comm. Rev.*, 29(4):251–262, 1999.

[43] Mahmassani H.S. Eisenman S.M. Fei, X. Sensor coverage and location for real-time traffic prediction in large-scale networks. *Transportation Research Record*, 2039(1):1–15, 2007.

[44] L. C. Freeman. A set of measures of centrality based on betweenness. *Sociometry*, 40(1):35–41, 1977.

[45] WARDROP J. G. Some theoretical aspects of road traffic research. *Proceedings of the Institution of Civil Engineers*, 1:325–378, 1952.

[46] B.P. Gerkey and M.J. Mataric. Sold! market methods for multi-robot control. *IEEE Transactions on Robotics and Automation, Special Issue on Multi-robot Systems*, 2002.

[47] Benjamin Gompertz. On the nature of the function expressive of the law of human mortality, and on a new mode of determining the value of life contingencies. *Philosophical Transactions of the Royal Society of London*, 115:513–583, 1825.

[48] T. Haynes and S. Sen. *Evolving Behavioral Strategies in Predators and Prey*, volume 1042 of *Lecture Notes in Computer Science*, chapter Adaptation and Learning in Multi-Agent Systems, pages 113–126. Springer, Berlin, 1986.

[49] Ian Henderson. Civilian intelligence agencies and the use of armed drones. In *Yearbook of International Humanitarian Law-2010*, pages 133–173. Springer, 2011.

[50] S. Hettiarachchi and W. Spears. Moving swarm formations through obstacle fields. In *International Conference on Artificial Intelligence*, 2005.

[51] P. Holme. Congestion and centrality in traffic flow on complex networks. *Advances in Complex Systems*, 6(2):163–176, 2003.

[52] Peeta S. Chu C.-H. Hu, S.-R. Identification of vehicle sensor locations for link-based network traffic applications. *Transportation Research Part B*, 43(8-9):873–894, 2009.

[53] Pavlos S. Kanaroglou, Michael Jerrett, Jason Morrison, Bernardo Beckerman, M. Altaf Arain, Nicolas L. Gilbert, and Jeffrey R. Brook. Establishing an air pollution monitoring network for intra-urban population exposure assessment: A location-allocation approach. *Atmospheric Environment*, 39(13):2399 – 2409, 2005. ¡ce:title¿12th International Symposium, Transport and Air Pollution¡/ce:title¿ ¡xocs:full-name¿12th International Symposium, Transport and Air Pollution¡/xocs:full-name¿.

[54] W. Kerr and D. Spears. Robotic simulation of gases for a surveillance task. In *Intelligent Robots and Systems (IROS 2005)*, pages 2905–2910, 2005.

[55] S. Kirkpatrick and J.J. Schneider. How smart does an agent need to be? *International Journal of Modern Physics*, C 16:139–155, 2005.

[56] B. Koopman. *Search and Screening: General Principles with Historical Applications*. Pergamon Press, 1980.

[57] B.O. Koopman. The theory of search ii, target detection. *Operations Research*, 4(5):503–531, 1956.

[58] R. E. Korf and W. Zhang. Performance of linear-space search algorithms. *Artificial Intelligence*, 79(2):241–292, 1995.

[59] Lo H. Lam, W. Accuracy of o-d estimates from traffic counts. *Traffic Engineering and Control*, 31:358–367, 1990.

[60] J. Lerner. *Network analysis: methodological foundations*, chapter Role Assignments. Springer LNCS 3418, 2005.

[61] Ouyang Y. Li, X. Reliable sensor deployment for network traffic surveillance. *Transportation Research Part B*, 45:218–231, 2011.

[62] F. Lorrain and H. C. White. Structural equivalence of individuals in social networks. *The Journal of Mathematical Sociology*, 1(1):49–80, 1971.

[63] M.J. Mataric. Designing emergent behaviors: From local interactions to collective intelligence. In J.Meyer, H.Roitblat, and S.Wilson, editors, *Proceedings of the Second International Conference on Simulation of Adaptive Behavior*, pages 432–441. MIT Press, 1992.

[64] P.M. Morse and G.E. Kimball. *Methods of operations research*. MIT Press and New York: Wiley, 1951.

[65] Alan T. Murray, Kamyoung Kim, James W. Davis, Raghu Machiraju, and Richard Parent. Coverage optimization to support security monitoring. *Computers, Environment and Urban Systems*, 31(2):133 – 147, 2007.

[66] K. Passino, M. Polycarpou, D. Jacques, M. Pachter, Y. Liu, Y. Yang, M. Flint, and M. Baum. *Cooperative Control for Autonomous Air Vehicles*, chapter Cooperative Control and Optimization. Kluwer Academic, Boston, 2002.

[67] Julian Tan Kok Ping, Ang Eng Ling, Tan Jun Quan, and Chua Yea Dat. Generic unmanned aerial vehicle (uav) for civilian application-a feasibility assessment and market survey on civilian application for aerial imaging. In *Sustainable Utilization and Development in Engineering and Technology (STUDENT), 2012 IEEE Conference on*, pages 289–294. IEEE, 2012.

[68] R. Puzis, Y. Elovici, and S. Dolev. Fast algorithm for successive computation of group betweenness centrality. *Phys. Rev. E*, 76(5):056709, 2007.

[69] R. Puzis, Y. Elovici, and S. Dolev. Finding the most prominent group in complex networks. *AI Comm.*, 20:287–296, 2007.

[70] R. Puzis, M. D. Klippel, Y. Elovici, and S. Dolev. Optimization of nids placement for protection of intercommunicating critical infrastructures. In *EuroISI*, 2007.

[71] Rami Puzis, Yaniv Altshuler, Yuval Elovici, Shlomo Bekhor, Yoram Shiftan, and Alex Pentland. Augmented betweenness centrality for environmentally-aware traffic monitoring in transportation networks. *Journal of Intelligent Transportation Systems*, 17(0):91–105, 2013.

[72] G. Rabideau, T. Estlin, T. Chien, and A. Barrett. A comparison of coordinated planning methods for cooperating rovers. In *Proceedings of the American Institute of Aeronautics and Astronautics (AIAA) Space Technology Conference*, 1999.

[73] Eyal Regev, Yaniv Altshuler, and Alfred M Bruckstein. The cooperative cleaners problem in stochastic dynamic environments. *arXiv preprint arXiv:1201.6322*, 2012.

[74] I. Rekleitis, V. Lee-Shuey, A. Peng Newz, and H. Choset. Limited communication, multi-robot team based coverage. In *IEEE International Conference on Robotics and Automation*, April 2004.

[75] Petri Rouvinen. Diffusion of digital mobile telephony: Are developing countries different? *Telecommunications Policy*, 30(1):46 – 63, 2006.

[76] Niklas Schörnig. Unmanned warfare: Towards a neo-interventionist era? In *The Armed Forces: Towards a Post-Interventionist Era?*, pages 221–235. Springer, 2013.

[77] J. Scott. *Social Network Analysis: A Handbook*. Sage Publications, London, 2000.

[78] B. Shucker and J.K. Bennett. Target tracking with distributed robotic macrosensors. In *Military Communications Conference (MILCOM 2005)*, volume 4, pages 2617–2623, 2005.

[79] L. Steels. Cooperation between distributed agents through self-organization. In Y.DeMazeau and J.P.Muller, editors, *Decentralized A.I - Proc. first European Workshop on Modeling Autonomous Agents in Multi-Agents world*, pages 175–196. Elsevier, 1990.

[80] R. Stern, R. Puzis, and A. Felner. Potential search: a bounded-cost search algorithm. In *AAAI 21st International Conference on Automated Planning and Scheduling (ICAPS)*, 2011.

[81] S. H. Strogatz. Exploring complex networks. *Nature*, 410:268–276, March 2001.

[82] S.M. Thayer, M.B. Dias, B.L. Digney, A. Stentz, B. Nabbe, and M. Hebert. Distributed robotic mapping of extreme environments. In *Proceedings of SPIE*, volume 4195 of *Mobile Robots XV and Telemanipulator and Telepresence Technologies VII*, 2000.

[83] A. Thorndike. Summary of antisubmarine warfare operations in world war ii. Summary report, NDRC Summary Report, 1946.

[84] P. Vincent and I. Rubin. A framework and analysis for cooperative search using uav swarms. In *ACM Simposium on applied computing*, 2004.

[85] I.A. Wagner, Y. Altshuler, V. Yanovski, and A.M. Bruckstein. Cooperative cleaners: A study in ant robotics. *The International Journal of Robotics Research (IJRR)*, 27(1):127–151, 2008.

[86] I.A. Wagner and A.M. Bruckstein. From ants to a(ge)nts: A special issue on ant—robotics. *Annals of Mathematics and Artificial Intelligence, Special Issue on Ant Robotics*, 31(1–4):1–6, 2001.

[87] S. Wasserman and K. Faust. *Social network analysis: Methods and applications*. Cambridge, England: Cambridge University Press., 1994.

[88] M.P. Wellman and P.R. Wurman. Market-aware agents for a multiagent world. *Robotics and Autonomous Systems*, 24:115–125, 1998.

[89] D. R. White and S. P. Borgatti. Betweenness centrality measures for directed graphs. *Social Networks*, 16:335–346, 1994.

[90] Yang C. Gan L. Yang, H. Models and algorithms for the screen line-based traffic-counting location problems. *Computers and Operations Research*, 33(3):836–858, 2006.

[91] Zhou J. Yang, H. Optimal traffic counting locations for origin-destination matrix estimation. *Transportation Research Part B*, 32(2):109–126, 1998.

[92] Charles Solomon-Leonid Kheifits Yehuda J. Gur, Shlomo Bekhor. Intercity person trip tables for nationwide transportation planning in israel obtained from massive cell phone data. *Transportation Research Record: Journal of the Transportation Research Board*, 2121:145–151, 2009.

[93] S.H. Yook, H. Jeong, and A.-L. Barabasi. Modeling the internet's large-scale topology. *Proceedings of the National Academy of Science*, 99(21):13382–13386, Oct. 2002.

[94] Shlomo Zilberstein. Using anytime algorithms in intelligent systems. *AI Magazine*, 17(3):73–83, 1996.